# Transforming Competition into Collaboration: The Revolutionary Role of Multi-Agent Systems and Language Models in Modern Organizations

Carlos Cruz

carlos.cruz@fgv.edu.br

carlos.jose.x.cruz@gmail.com

**Abstract**

This article explores the dynamic influence of computational entities based on multi-agent systems theory (SMA) combined with large language models (LLM), which are characterized by their ability to simulate complex human interactions, as a possibility to revolutionize human user interaction from the use of specialized artificial agents to support everything from operational organizational processes to strategic decision making based on applied knowledge and human orchestration. Previous investigations reveal that there are limitations, particularly in the autonomous approach of artificial agents, especially when dealing with new challenges and pragmatic tasks such as inducing logical reasoning and problem solving. It is also considered that traditional techniques, such as the stimulation of chains of thoughts, require explicit human guidance. In our approach we employ agents developed from large language models (LLM), each with distinct prototyping that considers behavioral elements, driven by strategies that stimulate the generation of knowledge based on the use case proposed in the scenario (role-play) business, using a discussion approach between agents (guided conversation). We demonstrate the potential of developing agents useful for organizational strategies, based on multi-agent system theories (SMA) and innovative uses based on large language models (LLM based), offering a differentiated and adaptable experiment to different applications, complexities, domains, and capabilities from LLM.

## 1 Introduction

Large Language Models (LLM) brought undeniable advancement to the area of Artificial Intelligence (AI), revolutionizing the software industry, especially in the field of Natural Language Processing (NLP).

LLMs have demonstrated extraordinary abilities beyond known text generation capabilities. One of the highly sophisticated capabilities is the ability of these models to systemically understand the underlying structures of written language, extending their usefulness to tasks ranging from comprehensible text generation to understanding programming code (Chen et al., 2021; Schick & Schütze, n.d.).

Studies highlight the exceptional proficiencies of LLMs in natural language processing (NLP) domains, which have the possibility of revolutionizing the potential of human-machine interaction using specialized virtual assistants (Rasal & Hauer, 2024) and supporting everything from operational processes to executive decision-making based on knowledge and human orchestration are redefining collaborative boundaries in organizations.

These abilities open a new path to exploring the dynamic influence of multi-agent systems (SMA), which can rely on artificial intelligence (AI) tools, specifically large language models (LLM), which offer extensive opportunities for exploration and applications. (Cheng et al., 2024) in modern organizational models, extracting high potential from Generative AI.

In our approach we employ agents developed from large language models (LLM), each with distinct prototyping that considers behavioral elements, driven by strategies that stimulate the generation of knowledge based on the use case proposed in the scenario (role-play). business, using a human-guided discussion approach between agents. Our approach demonstrates the potential for developing metahumans useful in organizational strategies, based on theories and principles from the field of study on multi-agent systems (SMA) combined with innovative uses based on large language models (LLM based), offering a differentiated experiment and adaptable to various organizational challenges in a controlled cyber environment. Since the different organizational models follow different structures, hierarchies, and management styles, they present complexities from the behavioral point of view of human beings and organizational culture (Langton, 2019). It is understood that with the evolution of AI, numerous aspects of organizational behavior are being changed, as well as perceptions, attitudes, levels of job satisfaction, motivation, the way groups and teams will interact and work, communication is one of the factors that will undergo changes important with the advent of AI above all, in addition to factors such as organizational culture and ethics.

SMAs or Multi-Agent Systems, a sub-area of Distributed Artificial Intelligence (IAD), play a fundamental role in the modernization of organizational processes and countless productive processes in human life. These systems are composed of autonomous agents, which can be software or robots, cooperating or competing to solve complex problems or perform tasks. They are particularly useful in environments where centralization of control and data processing is not viable or efficient (Wooldridge, 2009).

A practical example of the use of SMAs in organizations is in supply chain management. Here, different agents represent elements such as suppliers, distributors, and customers. They interact autonomously to optimize product flow, adjust production based on demand and minimize costs (Timm et al., 2003).

When integrated with large general-purpose language models (LLMs) such as GPT-3 and GPT-4 (OpenAI), Large (Mistral), Llama2 (Meta), and Gemini Pro (Google), SMAs gain an additional layer of capability analytical and decision-making. LLMs can process and analyze large volumes of textual data, which is crucial for understanding market trends, customer sentiment and making strategic decisions. This combination can be especially useful in areas such as customer service, where intelligent agents equipped with LLMs can provide quick and accurate responses, improving customer satisfaction (Brown et al., 2020).

Furthermore, integration between LLMs and SMAs in corporate environments can facilitate the analysis of unstructured data, such as emails, reports, and social media posts, allowing a more comprehensive and detailed view of the business environment. This can lead to more accurate insights and identification of emerging opportunities and risks (Devlin et al., 2019).

In this article we propose an LLM-based SMA approach as drivers of the development of intelligent agents that can act as meta-humans and these can be used for representations that can consider behaviors, functions, develop tasks with a high level of complexity and be, led and orchestrated by human beings, having the potential for customization and adaptability in different use cases that involve conducting organizational business internally and externally, with high scalability using the potential of the cloud, collaboration and dynamics where agents can collaborate in an adaptive way taking advantage of the machine learning capabilities, adaptation and use of LLMs parameters in real time.

Agents developed from large language models (LLMs) can interact with human actors at all organizational levels from operations to strategy, supporting executives in informed decision making, organizational learning, innovation, increasing creativity as well as reinforcing of organizational identity, seeking an internal balance between competition and collaboration between employees (Humans and meta-humans).

The proposed agents will be customized, refined and trained (not limited to the present study) based on the proposed representative use case, considering behavioral aspects that may touch on the concepts of "Jungian" psychoanalysis, where (1) archetypes that represent ideas, concepts or characters that have timeless appeal and resonate deeply with universal human experiences, however, archetypes do not have fixed or pre-defined forms (Jung, 2014a).

And the (2) personas, which is the psychic instance responsible for the interaction between the being and the community in general (Jung, 2014b), in theatrical or colloquial use, it is a social role or character played by an actor or persona developed with a focus on improving sales capabilities through the efficient design of the types of buyers of the organization's products and services based on the understanding of needs, desires and behaviors of customers to develop more targeted marketing strategies (Revella, 2015). Personas can also represent target users (Cooper et al., 2014), or as "fictitious, specific and concrete representations of target users" (J. Pruitt & T. Adlin, 2006).

According to (Cooper et al., 2014), A persona can represent a group of target users who share common behavioral characteristics, needs, and goals, written in the form of a detailed narrative about a specific, fictional person. These details make the persona feel like a real person in the designers' minds. Thus, by using a narrative, image, and name, a persona provides the human-computer interaction professional with a vivid, specific design that manufactures the persona to seem like a real person in the designers' minds (Miaskiewicz et al., 2008). The narrative also addresses the persona's goals, needs, and frustrations that are relevant to the product or system being designed (Maness et al., 2008). In (Norman, 2002) explains that in the context of personas, empathy is necessary to allow understanding and identification with the user population, to better ensure that they will be able to take advantage of the product or service.

Empathizing with personas allows the design team to stop talking about the general user when making product design decisions.

This profound shift from talking about users in general to understanding and identifying with personas'

needs and goals allows designers to more effectively address users' needs (Maness et al., 2008).

The fact of the relationship between archetype and persona is interesting from a practical point of view that considers studies and theories of Carl Gustav Jung's analytical psychology, since this relationship resides in the fact that the persona can be understood as an expression of the archetype of the "self". social" or "public self".

The persona is the way of presenting oneself to the world, and this choice can be influenced by archetypes present in the collective unconscious. For example, one might adopt a "hero" archetype in one's public life, trying to be courageous and resilient in all situations. In this case, the hero archetype is influencing the persona (extroverted reflective, introverted reflective, introverted perceptive, extroverted perceptive), as can also occur with archetypes such as "Explorer", "Rebel", "Creator", "Ruler", "Shadows" among others (Jung, 2014a).

New aspects of interaction create paradigms regarding business interactions (Huang & Rust, 2018). It is worth noting that unlike traditional and simplistic approaches that develop only individual agents, the strategy proposed in this article is dynamic, flexible and will seek to bring useful insights to the next generation of organizations driven by data, AI and multi-agent systems to transform competition into collaboration, bringing a discussion about how AI can increase and improve collective intelligence from the point of view of facilitating collaboration and human learning, considering the optimization and aggregation of information effectively (Malone, 2022).

Previous investigations reveal that there are limitations, particularly in the autonomous approach of agents to new challenges, pragmatic tasks such as inducing logical reasoning and problem solving, it is also considered that traditional techniques, such as stimulating chains of thoughts, require human guidance explicit (Devlin et al., 2019; Reed et al., 2022).

In this experiment, we will employ structured coding to orchestrate a simulated conversation between characters, with the purpose of reflecting, through text, the various attributes within a specific context and domain. When understanding these capabilities, it is critical to consider the ethical implications and risks associated with using Artificial Intelligence (AI), such as privacy issues, data security, and the possibility of algorithmic bias. Therefore, the implementation of these technologies demands a careful approach, with clear policies and protocols to ensure their responsible and ethical use (Russel & Norvig Peter, 2022).

Combining these concepts with the potential of AI enables the creative design of artificial agents, which can take the form of "meta-humans". These agents consider the integration of ideas, forms, knowledge, and specialties from the real world, such as AI, Computer Science, Software Engineering, Computational Linguistics, Data Science, Multi-Agent Systems Theory, Cognitive and Behavioral Psychology, Knowledge Management, Design, Organizational Modeling, Ethics in Artificial Intelligence, and use of Data.

In this case study, we will simulate the vision, knowledge and behaviors based on the role called Chief Financial Officer (CFO), an experienced persona in financial organizations. This persona will be personified as CFO (traditional) and CFO (Bold), performing the designated role-play, which may have defined strategic thinking alignments and in line with the organization's strategy and objectives.

In subsequent sections, we will delve deeper into aspects that demonstrate how LLMs and multi-agent systems can reshape the contemporary organizational landscape.

## 2   Methodology

### 2.1   General Approach

In this session we will cover the main components of the experimental platform that can support the development of this case study. This experiment will be developed using LLM (GPT-3.5), considering that each of the characters will have a distinct and representative prototype simulated from the proposed use case in business role-play scenarios, using a debate strategy to generate knowledge in your answers.

Note that the platform is not conditioned to the use only of OpenAI's LLM (GPT-3.5), which is flexible for the use of other general-purpose technologies such as GPT-4 (OpenAI), Large (Mistral), Llama2 (Meta) and Gemini Pro (Google).

The case study will be aimed at the **financial industry**, using a simulated challenge in a reduced format (business role play), the construction of the characters considers the role known as Chief Financial Officer (CFO). The research, **"In the face of volatility, CFOs—and their organizations—adapt"**, conducted by Mckinsey consultancy in 2023, was used as inspiration for developing the context of the case study.

To direct this and future studies, a descriptive macro of the components of the architectural framework of the experimental platform under development.

# Multiagent Data & AI Based Platform - Overview

The proposed framework (see figure 1) is a reference considering the components of (C1-C14) and will be used to refer to the various components and concepts used in this and future studies. When dealing with topics with great scope, such as multi-agent systems (SMA), Artificial Intelligence (AI), large language models (LLMs), the figure below (figure 1) is not intended to be an exhaustive model and will be referenced in the paragraphs below by its components (e.g. C1: Data; C4: Embedding Model).

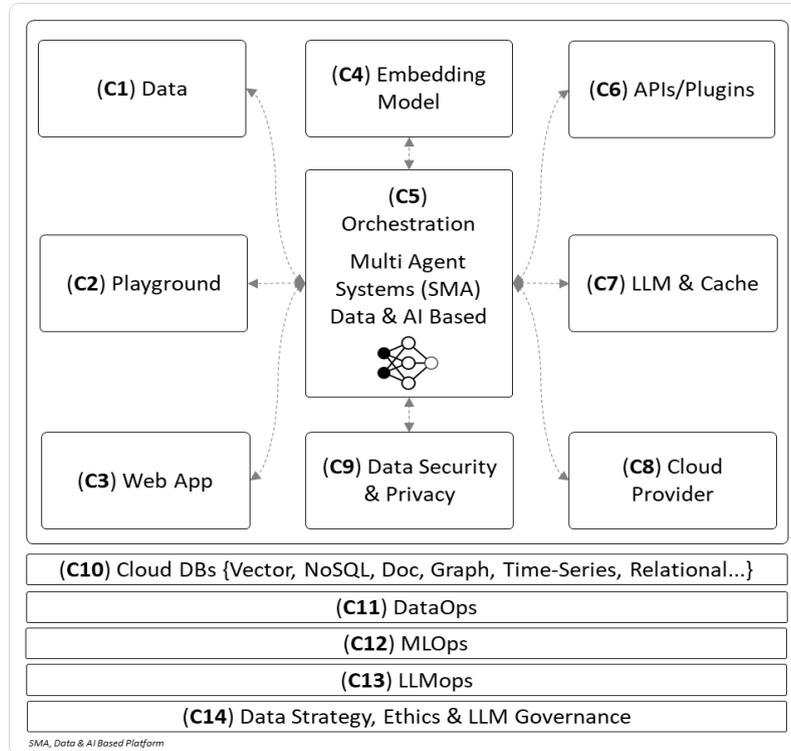

**Figure 1:** Representation of the main components of the framework

## 2.1.1   C1 – Data (General Point of View)

It is evident that a detailed understanding of the business processes, roles, and behaviors to be represented digitally are important, highlighting the need to understand the data inherent to the representations and development of the platform, which will be data driven.

The acquisition of classified and high-quality data are highly relevant factors for the orchestration of agents to occur effectively and close to the reality expected in a simulation.

Data quality is one of the main challenges when dealing with different formats and types of data, and these are essential requirements. Especially historical, contextual data, or special specifications that depend on the type of business process, persona, and function to be built, such data can prove to be complex as it can reside in different media such as proprietary platforms, third-party platforms, API's, varied formats and even image documents. How was the development of the BlombergGPT model (Wu et al., 2023; Yang et al., 2023), who achieved impressive results by capitalizing on exclusive access to specialized data to train large language models specifically in finance.

This highlights the need to use a data-centric approach, use of engineering protocols, data quality, and training requiring high expertise in engineering, and data governance. In this sense, we believe it is necessary to highlight the architectural layers prior to the proposed framework (see figure 1), as follows:

- **Data Source Layer:** This layer largely represents the data necessary to build the archetypes, personas, behaviors, demographic data, psychographic characteristics, behaviors, needs, objectives, experiences, knowledge, past, professional environment when it comes to the metahuman or agent. Otherwise, this is also the layer that will serve as input to capture data about the business model and processes inherent to each industry.

- **Data Engineering Layer:** This layer addresses the challenge of processing times (extraction,

transformation, and loading), data validity and sensitivity. There are data that have high temporal sensitivity and low signal-to-noise ratio. (Fama, 1970), especially when considering information from the financial industry and market efficiency.

### 2.1.2 C1 - Data (Financial Services Point of View)

Considering the proposed use case and the specialty of the personas, who are experts in the financial sector, it is important to highlight the unique characteristics that the data used in this industry presents. In this way, our study considers a data-centric approach, specifically highlighting the application industry.

Financial sector data originates from a wide variety of sources, types of operations, types of transactions, various legacy systems, with unique characteristics such as financial news, discussions originating from social media, various announcements about indicators, financial products, among many others. These data sources have different characteristics such as:

- **Time**: Much information is updated opportunistically (as is the case with seasonal and specific reports) and has a short validity window as it is captured at moments of interest.
- **Dynamism**: News and indicators change quickly in response to economic stimuli, unexpected movements conditioning the responses of investors, organizations and above all information.
- **Influence**: Information and news have a significant impact on decision-making, finance, and financial markets that can lead to sudden movements across the entire financial market, especially in assets related to specific investments and products.
- **Granularity**: Data can be granular into larger and smaller grain of data (Kimball & Ross, 2015), relating to the financial situation of an organization, including assets, liabilities, revenue, margin, profitability, forecasts, etc.

**Company Financial Records** are official documents that can be open to the public, submitted to regulators, which can be used to obtain insights into financial success, strategy and these can represent important information to define:

- **Reliability**: Different documents offer information at different granular levels, financial status, asset value, revenue, margin, profitability.
- **Periodicidade**: São informações periódicas para acompanhamento da situação financeira e podem ser mensais, semanais, anuais, quartil)
- **Impact**: There are various announcements and publications that can provide information about important impacts on the organization's financial health and that can influence the market, share prices, investor, and consumer views.

**Social media**: Since economics is not just an exact discipline, social media can reflect different views based on feelings that are not mathematically based and can bring variables such as:

- **Variability and tensions**: Texts described on social networks (discussions and posts) are highly variable in tone, content, quality, and this brings enormous complexities when adopted as a valid source of information for certain analyzes.
- **Real-time feelings and visions**: Social media can be excellent platforms for capturing real-time data that can refer to market "sentiment" over time and can express the "trend" and/or changes in "opinion" of users.
- **Volatility**: The data expressed on social networks is highly volatile and notably can be intentionally influenced and quickly change responses to events (news and facts), bringing often unexpected movements to the market.
- **Trends**: These are the responses to market movements that are observed in the various communication channels typically used by professionals in the financial sector such as Google, Financial Teams among many others that may have critical importance on financial market movements.
- **Perspective**: It is the combination of different market perspectives considering the different experts who analyze the data and can give investment advice, organizational movements that can go in different directions and depend on the informational context of the market.

- **Market Sentiment**: Writing on these platforms can reflect the feeling of a "collective" about specific topics and sectors or about the market in general, providing valuable information for decision-making by organizations and people.
  - **Coverage range**: It is about the scope of information that can cover a product, service, market segment in a specific way, limiting analyzes or the opposite.

It is interesting and challenging to develop agents that can deal with this information where each of the data sources, when combined, can provide exclusive insights into the financial world, and bring handling complexities (high temporal sensitivity, high dynamism and low signal-to- noise (SNR). By integrating different types of data, properly understood, classified, and qualified, agents can bring understandings, points of view, new questions, and comprehensive answers to support modern organizations in the effective use of data and allow learning both for taking decision-making and executive education.

### 2.1.3 C2 - Playground

Represents the integration and/or use of web-based development platforms (OpenAI, HuggingChat, Cohere) to experiment, test, deploy code, prompts and/or machine learning models using large language models (LLM), designed to accelerate the process of writing, testing and code generation.

### 2.1.4 C3 - Web App

It is the WEB-based application layer and/or software hosted on a cloud provider (e.g. Azure Cloud, Amazon Web Services, Google Cloud Platform), which demonstrates the potential capacity of using the proposed framework considering the potential of large data models. language (LLM) directed to this channel.

### 2.1.5 C4 - Embedding Models

These are models that can map words or phrases to vectors in a continuous space (these models are referenced at coding time), so that semantically similar words are represented by nearby vectors. They consist of "tokens" that are treated as resources in your dataset. These models are trained on large linguistic datasets to capture contextual and semantic relationships with for example Word2Vec, GloVe (Global Vectors for Word Representation), BERT (Bidirectional Encoder Representations from Transformers), OpenAI Ada Embedding 2, Cohere AI.

### 2.1.6 C5 - Orchestration (SMA, Data & AI Based)

In scenarios involving one or multiple agents (Data & AI based) there is a need to orchestrate interactions between digital agents combined with human interactions. This orchestration component represents a fundamental functionality in managing interactions and coordinating data flow. It is worth mentioning that the architecture of multi-agent systems refers to the multiple use of data and AI including specialist LLMs, and this means that they have their own resources such as (embeddings, dictionaries, related tables, algorithms, etc.). Below functions provided (not exhaustive) in the orchestration module with multiple agents:

- **Command Interpretation:**
  - The orchestration component must be capable of interpreting commands given by humans to start, pause, resume, or end activities. This would involve analyzing verbalized (use of voice) and textual (use of keyboard) instructions.

- **Scripting:**
  - Based on the instructions, the orchestration component scripts actions, determining the execution rules, topics, actions, processes, and functionalities to be executed by the agents.

- **Shift Control:**
  - Ensures a fair and efficient distribution of speaking turns among agents. It could also manage each agent's speaking time to keep communication balanced.

- **Integration (Systems, Data, Multi-Agent LLMs/Meta-humans…):**
  - Integration, communication with agents to transmit instructions, collect responses and coordinate communication, organize, and govern the information and knowledge generated.

- **Performance Monitoring:**
  - Monitors the performance of agents, evaluating relevance, coherence, and respect for the rules of the debate. This may involve analyzing specific metrics generated by language models.
- **Generation of Prompts, Questions and Stimulus:**
  - Based on the context in the instructions, the orchestration component generates questions, stimuli or challenges aimed at specific agents to stimulate discussions, propose new ideas, new lines of thought to obtain creative forms of collaboration and contribution with human beings.
- **Human Intervention:**
- Ability for real-time human intervention to correct misunderstandings, adjust the script or add additional information for agents to use.
- **Evaluation and feedback:**
  - After interactions, the orchestration component provides a summary of agents' performance, highlighting strengths and areas for improvement.

### 2.1.7 C6 - APIs/Plugins

It is the interface that allows interaction between different software and in the context of LLMs, APIs can be used to integrate language models into applications, websites, or services such as OpenAI provides an API to access GPT models, allowing developers to easily incorporate the Text generation in your applications. Plugins represent additional modules that extend the functionalities of an application and, in the case of LLMs, can be used to integrate Generative AI functionalities into different types of applications.

### 2.1.8 C7 - LLM & Cache

It is a component developed to enable a memory manipulation strategy in certain scenarios to temporarily store intermediate representations of words or phrases for faster access to previously processed information, depending on the type of application.

### 2.1.9 C8 - Cloud Provider

It is the cloud computing platform that can offer global hosting services for solutions such as Google Cloud Platform, Amazon Web Services or Microsoft Azure.

### 2.1.10 C9 – Data Security & Privacy

It represents the layer and technological mechanisms responsible for the security and privacy of the data used in each use proposition, in accordance with the rules and definitions in (C13) (see Figure 1).

### 2.1.11 C10 - Cloud DBs

It represents the usage layer of different types of databases in the cloud, they are data storage services that operate to meet different data requirements and have different uses like (store data, enable RAG and/ or fine-tunning, Grounding, support data governance strategies, manage data, knowledge, cache, and metadata), when it comes to LLM based solutions.

- **Relational databases:** Store data in related tables and can be used to store metadata, train models, and maintain structured information about the language and use of solutions.
- **NoSQL databases:** They do not follow the previous model, they can be oriented by documents, key-value, column, or graphs and are useful for storing large volumes of unstructured data such as long texts, unprocessed texts and can be used for model training.
- **Document database:** Store data in the form of graphs, with "nodes" representing entities and edges representing relationships and can be used to represent complex semantic relationships, such as the knowledge structure used in some LLMs.
- **Time Series Databases:** They are designed to store and retrieve data organized by time stamp or time interval and can be useful for storing temporal data that may reflect language changes over time.

- **Vector Databases:** Focused on storing and processing vector data, such as distributed representations of words and can be used to store pre-trained embeddings, facilitating quick access to semantic representations of words.

### 2.1.12 C11 – DataOps

These are the practices, cultures and methods that aim to integrate data, improve collaboration and productivity between teams related to data to supply the different models (ML, DL, GenAI) and in turn the LLMs, being important for management of the data lifecycle.

Typically, the team that makes up the DataOps team are data scientists, data engineers, analysts, among others. The effectiveness, scalability and quality of the models depend on the DataOps strategy implemented. In the context of use for operations based on LLMs DataOps can:

- **Increase collaborative efficiency:** Incorporates practices that promote efficient collaboration between data teams, from the acquisition to the implementation of language models, such as the establishment of processes, communication, knowledge sharing and integration between team members.
- **Structuring data management and lifecycle:** Involves effective management of the data lifecycle, from collection and preparation to training and model deployment. Promotes good practices for configurations, implementations, automations (cleaning, transformations) of linguistic data for training models in data pipelines.
- **Workflow automation:** Automation of repetitive processes to improve operational efficiency and reduce human errors and can be used to use automation tools to automate the execution of model training experiments, hyperparameter tuning and performance evaluation.
- **Versioning of data and models:** Highly useful for versioning data and models to track changes, facilitating the reproduction of experiments and ensuring consistency, they can be used in version control systems to track changes in training data sets and models, allowing a consistent analysis of results.
- **Continuous deployment:** DataOps brings practices that enable the efficient and continuous deployment of models in production environments. Use techniques such as deploying models in containers to facilitate continuous integration and continuous delivery (CI/CD) of language models.
- **Performance monitoring:** Implementing systems to continuously monitor (after establishing metrics, alerts, and notifications) the performance of models in production is one of the practices foreseen in DataOps.
- **Security and compliance:** DataOps provides a guide for adopting practices that guarantee data security and adherence to compliance and regulations, such as the implementation of security measures with encryption, access control and auditing to protect sensitive data processed by the models.

### 2.1.13 C12 – MLOps

MLOps is an ML culture and practice that unifies ML application development (Dev) with ML system deployment and operations (Ops). Your organization can use MLOps to automate and standardize processes across the entire ML lifecycle, broadly integrated with DataOps practices.

### 2.1.14 C13 – LLMOps

LLMops is not a very common term, and its definitions may be connected to the concept of DataOps applied to LLMs (see item 2.1.12). LLMops allows the implementation, monitoring, and adjustments of LLMs, which require collaboration between DataOps, data science, DevOps teams, like the strategies adopted in MLOps, widely integrated with DataOps practices.

### 2.1.15 C14 - Data and LLM Governance

Data governance refers to practices and policies to ensure the quality, integrity, security, and compliance of data in an organization. It involves defining guidelines, responsibilities, and processes to effectively manage the data used in building the solution (organizational level) and involves the points (not exhaustive):

- **Ethics in Language:**
  - Establish ethical guidelines to ensure that language models do not promote bias, hate speech, or discrimination.
- **Interpretability and Transparency:**
  - Require models to be interpretable and transparent in their decisions, especially in critical applications.
- **Responsible Training:**
  - Ensure that model training is carried out responsibly, avoiding bias and promoting equity.
- **Continuous Monitoring:**
  - Implement monitoring systems to detect changes in model performance or behavior over time.
- **Impact Assessment:**
  - Conduct impact assessments to understand how models affect users and communities and adjust as necessary.
- **Data Privacy:**
  - Adopt measures to protect the privacy of data used in training and ensure compliance with privacy regulations.

The proposed structure (see figure 1) aims to demonstrate the potential of a platform developed to build artificial minds, based on the theories, principles and techniques presented, demonstrating that human beings can orchestrate the agents, personified in metahumans while maintains knowledge and control and governance possibilities applied to different use cases.

## 2.2 Development Strategy

The development strategy is described below, (see Figure 2) which schematically demonstrates the flow that will be used to enable the "conversation" between the two declared artificial **John**, classic CFO (Focus on capital protection) and **Anne**, Bold CFO (Focus on Growth).

As shown in figure 2, in (01) the context considering the economic scenario is inserted via code (see Item 2.2.1). In (02) we inserted via code an open question focused on generating knowledge and encouraging discussion between the two agents. In (3) there is an open discussion that takes place in 50 rounds (questions and discursive answers). In (4) there is a qualitative and quantitative evaluation of the transcription of the conversation between agents 1 and 2.

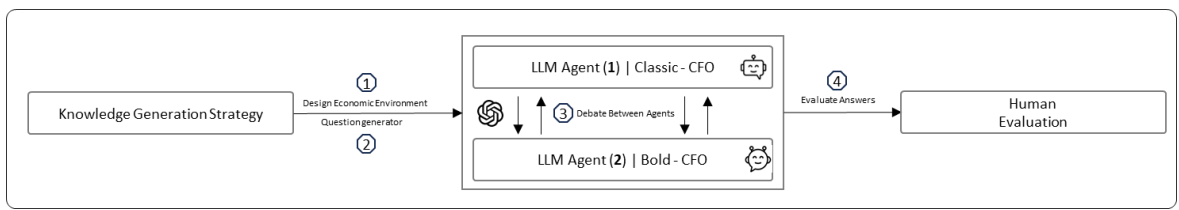

**Figure 2** - Schematic representation of the data flow and actions of each actor

We will deliberately use OpenAI's GTP-3.5 technology to illustrate the developments by combining free-to-use data (web), code developments using the OpenAI playground, use of the Azure cloud manager and components (KeyVault), API (OpenAI on Azure) and Python programming language. The GPT-3.5 parameters that are used in this experiment such as vocabulary size, embedding size, temperature, top_p, max_tokens, presence_penalty, Frequency_penalty combined with the context prompt engineering techniques for generating knowledge, resources, configurations, definitions are described in the sessions (2.2.3) and (2.2.4) and in the code distributed and highlighted in (2.2.5).

### 2.2.1 General Business Context

We understand that the modern business environment imposes significant challenges on CFOs, requiring rapid adaptations to technological, regulatory, and economic changes. The increasing complexity of commercial operations and the need to make quick strategic decisions in search of opportunities and competitive advantage have become fundamental aspects for the role of the CFO of financial organizations, highlighting the critical importance of financial risk management in operations (Saunders, 2018). And the use of robust analytical approach (Davenport, 2007), to obtain competitive advantage taking into account generic positioning strategies (Porter, 1980).

For this simulation we will highlight the general business context of the scenario:

- **Financial Market Volatility:** Global economic uncertainty and changes in financial markets introduce significant risks that directly affect financial decisions.

- **Intense Competition:** Modern banks, often fintechs, present a competitive threat, requiring CFOs of traditional institutions to seek innovations to remain relevant.

- **Complex Regulations:** Increasing regulatory complexity poses compliance challenges and requires significant investments in systems and processes.

- **Efficient Cost Management:** The constant pressure for efficiency and cost reduction requires a strategic approach to financial management.

- **The use of data and Artificial Intelligence:** The strategic use of data has become an imperative for CFOs. Collecting, analyzing, and interpreting large data sets can provide valuable insights to guide financial decisions. Artificial intelligence, through advanced algorithms, can offer more accurate predictions, identify patterns, and automate processes, alleviating the operational burden.

The question that **Anne** (Bold CFO) will use to start the conversation with **John** (Classic CFO) is described below.

> *"We, CFOs, are having difficulty adapting to highly volatile economic conditions. But it is important to look for innovative ways to maintain growth and profitability. Do you agree?"*

Anne's question directs towards inducing logical reasoning and chain of thought to generate knowledge considering CFOs' ability to manage risks, make strategic decisions and implement operational changes to ensure business success in a challenging environment.

### 2.2.2 The Artificial Agents

In this experiment, the imagined archetypes will be based on the role of executives in the real world, called Classic CFO and Bold CFO, which do not necessarily represent archetypes in themselves, however in this context we will respectively treat this role as being the "guardian" of the organization's finances. and the "Rebel and revolutionary" who intends to grow the organization's capital.

The guardian and the revolutionary rebel are archetypes (Jung, 2014a). Let's join the two executives (agents) with different styles in a lively conversation about business challenges based on the topic below.

> *"How CFOs are balancing the need to adapt to volatile economic conditions with maintaining the growth and profitability of their organizations."*

**The macro characteristics of our artificial agents are as follows:**

**Characteristics of the Traditional CFO**

Jonh, Chief Financial Officer (CFO) Classic, will act as Chief Financial Officer (CFO) with experience in large and traditional financial institutions and has characteristics such as:

- **Financial Conservatism:** Tends to adopt a more conservative stance in relation to financial risks. Prioritizes stability and security in investment decisions.

- **Focus on Compliance:** Places great importance on compliance with accounting regulations and standards. Prioritizes legal compliance and seeks to minimize the risks associated with possible penalties.

- **Conventional Approach to Cost Management:** In cost management, the traditional CFO adopts a more conventional approach, seeking efficiency and expense reduction, but generally with a greater emphasis on stability than on innovation.

- **History-Based Decision Making:** Analyzing historical data is fundamental for the traditional CFO when making decisions. Experience and historical patterns are strong influencers in the formulation of financial strategies.

**Characteristics of the Bold CFO**

Anne, Chief Financial Officer (CFO) Capital Rise, will act as Chief Financial Officer (CFO) with experience in modern financial institutions (Neo Banks) and Fintechs, her characteristics are:

- **Risk Propensity and Innovation:** The bold CFO is more inclined to take calculated risks in search of growth opportunities. Willing to explore new strategies and innovations to drive financial performance.

- **Emphasis on Growth Strategies:** Unlike the traditional CFO, the bold CFO places a strong emphasis on growth strategies, including more aggressive investments in new markets, emerging technologies, and significant expansions.

- **Adoption of Disruptive Technologies:** The bold CFO is more open to incorporating disruptive technologies, such as artificial intelligence and blockchain, to improve operational efficiency and create competitive advantages.

- **Greater Flexibility in Cost Strategies:** In cost management, the bold CFO can adopt a more flexible approach, prioritizing investments that promote innovation and growth, even if this involves a temporary increase in costs.

- **Intensive Use of Real-Time Data:** Unlike the traditional CFO, the bold CFO intensively uses real-time data to guide decisions. Predictive analytics and dynamic data interpretation play a crucial role in your strategies.

Both the characteristics of the artificial agents John and Anne are illustrated in the session table (2.2.4 personas parameters).

**Note:** *The development of the characteristics of the artificial agents mentioned only aims to demonstrate the possibilities of representing personas according to their imagined characteristics. Therefore, it is important to be aware that when developing solutions that involve Artificial Intelligence, it is important to consider measures to avoid cognitive bias and prejudices and any type.*

### 2.2.3 GPT Parameters

The parameters below were distributed equally for both agents, using gpt-3.5.

Table 1 – Global GPT-3.5 Parameters

| Parameter | Type of parameter | Value |
|---|---|---|
| *presence penalty* | GPT Parameter (0.0 - 2) | 0.8 |
| *temperature* | GPT Parameter (0.0 - 2) | 0.8 |
| *top-p (Probability Mass)* | GPT Parameter (0.0 - 2) | 0.8 |
| *frequence Penalty* | GPT Parameter (0.0 - 2) | 0.8 |
| *max_tokens* | GPT Parameter (0.0 - 4096) | 100 |

### 2.2.4 Personas Parameters

The parameters below are part of the contextual and behavioral configuration of each persona.

Table 2 – Persona parameters

| Parameter | Type of parameter | Classic CFO | Bold CFO |
|---|---|---|---|
| *Adaptability to Change* | Persona Parameter (0.0 - 1) | 0.5 | 1 |
| *Argumentative Style* | Persona Parameter (0.0 - 1) | 0.8 | 0.8 |
| *Cautiousness in Speculative Scenarios* | Persona Parameter (0.0 - 1) | 1 | 0.7 |
| Conventional Approach to Cost Management | Persona Parameter (0.0 - 1) | 1 | 0.5 |
| Dynamic Context Awareness | Persona Parameter (0.0 - 1) | 0.8 | 0.8 |
| Emphasis on Short-Term Strategies | Persona Parameter (0.0 - 1) | 0.5 | 0.9 |
| Financial Conservatism | Persona Parameter (0.0 - 1) | 1 | 0.5 |
| Focus on Compliance | Persona Parameter (0.0 - 1) | 0.9 | 0.5 |
| Focus on Long-Term Strategy | Persona Parameter (0.0 - 1) | 0.8 | 0.5 |
| Growth Strategies | Persona Parameter (0.0 - 1) | 0.5 | 0.9 |
| History-Based Decision Making | Persona Parameter (0.0 - 1) | 1 | 0.5 |
| Inclusion of Case Studies | Persona Parameter (0.0 - 1) | 1 | 0.7 |
| Incorporation of Informal Language | Persona Parameter (0.0 - 1) | 0.0 | 0.8 |
| Opening to Speculative Scenarios | Persona Parameter (0.0 - 1) | 0.8 | 1 |
| Penalty for Absence of Risks | Persona Parameter (0.0 - 1) | 0.3 | 0.3 |
| Response Length | Persona Parameter (0.0 - 1) | 0.5 | 0.5 |
| Risk Propensity | Persona Parameter (0.0 - 1) | 0.5 | 0.9 |
| Role-play Directive | Persona Parameter (0.0 - 1) | 0.8 | 0.8 |
| Role-play Driven by Innovations | Persona Parameter (0.0 - 1) | 0.8 | 1 |
| Sensitivity to Financial Sector Trends | Persona Parameter (0.0 - 1) | 0.8 | 0.7 |
| Sustainability Consideration | Persona Parameter (0.0 - 1) | 0.7 | 0.7 |
| Technology innovation | Persona Parameter (0.0 - 1) | 0.7 | 1 |
| Use of Financial Terminology | Persona Parameter (0.0 - 1) | 1 | 1 |
| Use of Proactive Language | Persona Parameter (0.0 - 1) | 0.8 | 0.8 |
| Weighting of Certain Keywords | Persona Parameter (0.0 - 1) | 0.6 | 0.6 |
| Intensive Use of Real-Time Data | Persona Parameter (0.0 - 1) | 0.6 | 1 |
| Logical and Reasoning | Persona Parameter (0.0 - 1) | 0.9 | 0.6 |
| Formal Language Tone | Persona Parameter (0.0 - 1) | 1 | 0.7 |
| Casual Language Tone | Persona Parameter (0.0 - 1) | 0.0 | 0.0 |

### 2.2.5 Code Overview

The code (Python) is fully available under the MIT license, (https://github.com/carlosXcruzCode/Compet_Colab_SMA_LLM) presents highlighted parts of the code (see table 3) that contribute to the simulation of the guided conversation between the two artificial agents (not integral).

**Table 3 -** Code Snippets

| Main Features | Description | Python Code |
|---|---|---|
| Openai API Configuration | Configuration of the OpenAI API, including obtaining the API key either from environment variables or Azure Key Vault. | python API_KEY = os.environ.get("OPENAI_API_KEY")<br>if API_KEY is None:<br>   ...<br>openai.api_key = API_KEY |
| Prompt Generation And Response | Functions for creating prompts that initialize the conversation and respond to previous interactions. Utilizes OpenAI's engine to interact with the GPT-3.5 model, generating responses based on provided prompts. | python def initialize_conversation(topic='', character=''):<br>   ...<br>def respond_prompt(response, topic='', character=''):<br>   ...<br>def openai_request(instructions, task, model_engine='gpt-3.5-turbo'):<br>   ... |
| *Character Descriptions* | Detailed descriptions and attributes of the characters involved in the conversation, including their professional characteristics and attributes influencing their behavior in the conversation. | python character_1 = { ... }<br>character_2 = { ... }<br> <br>if i % 2 == 0:<br>   ...<br>else:<br>   ... |
| *Conversation Loop* | Manages the conversation rounds and alternates between prompts of different characters based on the round number. | python for i in range(conversation_rounds):<br>   ... |
| *Conversation Storage* | Handles the storage of conversation logs in HTML files with timestamps, ensuring preservation of conversation history. | python filename = f'{path}/GPTconversation_{timestamp}.html'<br>with open(filename, 'w') as f:<br>   f.write(conversation) |
| *Waiting Between Responses* | Implements a waiting period between responses to simulate a more natural conversation flow. | python time.sleep(15) |
| *Error Handling* | Deals with potential errors or exceptions that may occur during the execution of the code. | python try:<br>   ...<br>except Exception as e:<br>   print(f"An error occurred: {str(e)}") |
| *Security In Api Key Management* | Secure management of the API key by first checking environment variables and then attempting to retrieve it from Azure Key Vault. | python if API_KEY is None:<br>   ...<br>client = SecretClient(f"https://{keyvault_name}.vault.azure.net/", AzureCliCredential())<br>API_KEY = client.get_secret('Your OPENAI_API_KEY').value |
| *Flexible Openai Model Configuration* | Flexibility to configure the OpenAI model used for text generation, allowing users to select the most suitable model for their application needs. | python def openai_request(instructions, task, model_engine='gpt-3.5-turbo'):<br>   ... |
| *Personalization Of Conversation Characters* | Personalizes conversation characters with detailed descriptions and attributes, influencing their interaction style. | python character_1 = { ... }<br>character_2 = { ... } |
| *Html Conversation Logging* | Logs conversations into HTML files with timestamps for easy access and reference to conversation history. | python with open(filename, 'w') as f:<br>   f.write(conversation) |
| *Proper Error Handling* | Ensures proper handling of errors or exceptions that may occur during code execution, enhancing overall robustness and reliability of the system. | python except Exception as e:<br>   print(f"An error occurred: {str(e)}") |

## 3 Discussion and Results

The integration of computational entities based on the principles of multi-agent systems (SMA) highlighting the autonomy, interdependence, coordination, modeling, simulation, and distribution possibilities of agents (Wooldridge, 2009).

Large language models (LLM) that have capabilities for executing multiple tasks are revolutionary in the field of natural language processing (NLP), since these models are trained in an unsupervised way on several tasks. This approach allows models to learn rich, generalized language representations (Brown et al., 2020).

These combined features offer a new paradigm for human-machine interaction in the financial industry. Our approach focuses on utilizing specialized artificial to support a wide range of organizational processes, from operational to strategic decision making.

The results of our research demonstrate that this approach has the potential to revolutionize the way organizations conduct their activities, providing greater efficiency, precision, and adaptability.

One of the main contributions of our study is the ability of artificial developed from large language models to simulate complex human interactions. By considering behavioral elements and directing strategies that stimulate the generation of knowledge, our agents can perform advanced cognitive tasks, such as problem solving and logical reasoning.

This suggests that although there are limitations in the autonomous approach of artificial, the combination of SMA with LLM can overcome these limitations, offering a new horizon for human-machine interaction.

Furthermore, the use of discussion approaches between agents, such as guided conversation, has proven to be effective in generating knowledge from simulated business scenarios.

The use of role-play in a controlled cyber environment provides a dynamic and adaptive means to explore diverse organizational situations, allowing agents to learn and evolve in real time.

Our results also highlight the potential for developing "metahumans" useful in organizational strategies. By empowering artificial with advanced cognitive skills and guiding them through adaptive strategies, our approach offers an innovative solution to organizational challenges in an ever-changing environment, particularly from an organizational learning perspective.

Learning from observing the discussion between actors with complementary views, where interested parties can maintain control, with a certain distance while learning. This transforms the analysis of a discussion into an attractive way of obtaining knowledge by bringing diversity of perspectives where in a discussion, different points of view and experiences are shared, especially considering areas of knowledge with high similarity as is the case of this study, the which can provide a broader and deeper insight into a given subject. This allows participants to consider aspects that may not have been previously considered.

Just like in the real world, encouraging the exchange of ideas and opposing arguments, discussion facilitates the exchange of ideas between participants, allowing new concepts to be presented and debated. This process can lead to creative insights and innovative solutions to complex problems, identifying strengths and franks of each position, enabling the identification of sometimes non-trivial knowledge.

Highly relevant point is the aspect of collaborative learning, which promotes collaborative learning, where participants can benefit from each other, building on each other's ideas and correcting mistaken understandings. This creates a dynamic and interactive learning environment seeking practical applicability for each organization.

Considering the theme, the use of LLM to simulate discussion between agents, personified, we highlight the following analyses:

### Qualitative Analysis of The Conversation Between Artificial Agents

The transcription of the conversation between the agents presents a structured discussion between two personas representing different approaches in the financial area: John, the classic CFO, and Anne, the bold CFO.

They debate issues such as sustainability, technological innovation, adapting to volatile economic conditions and the use of real-time data for decision-making. John tends to emphasize the importance of stability, security, and historical data analysis, while Anne highlights the need for adaptability, innovation, and use of real-time data.

The conversation reflects different perspectives and approaches to the challenges faced by CFOs in the financial industry.

The following table provides a (non-exhaustive) summary of the discussion between the agents, reinforcing the characteristics shaped by the design (code, parameters, context, prompt engineering) of each persona.

**Table 4 -** Conversational Analysis (Summary)

| Persona | Excerpts Of Conversation | Persona Characteristics |
|---|---|---|
| Anne | "Absolutely, sustainability initiatives can also lead to cost savings through energy efficiency and waste reduction, further enhancing the financial performance of the organization." | Emphasizes the importance of sustainability and financial efficiency. |
| John | "Indeed, implementing technological innovations can help streamline processes, improve efficiency, and reduce costs in the long term." | Highlights the importance of technological innovation to reduce costs. |
| Anne | "Absolutely, leveraging disruptive technologies like artificial intelligence and blockchain can provide a competitive edge and enhance operational efficiency." | Values the adoption of disruptive technologies for competitive advantage. |
| John | "Indeed, investing in sustainable practices and cost-effective measures can help mitigate risks and ensure long-term profitability despite economic volatility." | Prioritizes sustainable practices and cost-effective measures to mitigate risks. |

| | Anne | *"Absolutely, real-time data allows CFOs to quickly identify trends and opportunities, enabling them to adapt their strategies accordingly in volatile economic environments."* | *Highlights the importance of using real-time data for strategic adaptation.* |
|---|---|---|---|

The following table provides a **frequency analysis**, considering each persona, behavior, number of lines, main words used that reinforce the positioning of the characters that can be contrasted with the parameters, context and descriptions entered via code.

**Table 5 -** Frequency Analysis (Summary)

| Persona | Behavior | Rounds | Keywords | Frequency Analysis |
|---|---|---|---|---|
| *Anne* | Boldness | 30 | sustainability, disruptive technologies, AI, blockchain, real-time data, predictive analytics, growth, innovation | High frequency of words related to boldness and innovation, such as "disruptive technologies", "real-time data", "predictive analytics". |
| *John* | Conservatism | 30 | sustainability, historical data, stability, security, conservative financial approach | High frequency of words related to conservatism and stability, such as "historical data", "stability", "conservative financial approach". |

Analysis of the characters' speeches reveals a positive feeling with both personas demonstrating agreement and enthusiasm regarding topics such as sustainability, innovation and adaptation to volatile economic conditions, topics discussed include sustainability, technological innovation, use of real-time data, adaptation to volatile economic conditions and strategies to ensure long-term growth and profitability.

Keywords in the discussion often include sustainability, innovation, efficiency, real-time data, economic volatility, long-term strategies, and profitability, with thematic coherence in each persona's speeches, with Anne emphasizing the importance of innovation and rapid adaptation, while John highlights the need for stability and historical data analysis.

Considering convergence and divergence there are areas of convergence between John and Anne's views, especially in relation to the importance of sustainability and technological innovation. However, there are also divergences regarding approaches to dealing with economic volatility and the use of real-time data.

The results of our investigation revealed that the dynamic integration of SMA and LLM-based computational entities provides significant potential for improving human-machine interaction. Our artificial have demonstrated skills that can be refined and generate knowledge in different areas, applied in different industries in a few diverse use cases.

## 4 Limitations

The revolutionary potential that multi-agent systems (SMA), based on large language models (LLM-Based) possess, is undeniable and opens doors to creating value in infinite use cases.

However, given the need for realistic simulation of real-world processes or systems, it is important to recognize the (non-exhaustive) and unavoidable limitations that exist in this experimental study, as it is the first in a series considering the development of the platform mentioned in this article. document.

**Limited Generalization:** The case study is focused on a specific industry (finance) and specific challenges faced by CFOs. This may limit the generalizability of results to other industries or to different functions within organizations.

**Character Simplification:** Characters are simplified into archetypes (Classic CFO and Bold CFO), which may not fully capture the diversity of profiles found in real CFOs. The complexity of human behavior and interactions between executives can be difficult to fully represent in a simulated scenario.

**The simplified architecture of the agents:** It is still not enough to deal with highly complex tasks considering the level of human cognition, although evolution and adaptation for use in different conditions and use cases is possible, since the platform aims to become a platform for orchestrating multi-agent systems (SMA), Multi-LLMs (LLM-based), has long been developed, especially when dealing with multidimensional environments that interact with different data platforms, different applications, different cloud providers and multifactor parameterization to simulate increasingly complex archetypes, personas, behaviors and activities.

**Dependence on Language Models:** The effectiveness of the simulation depends on the ability of language models, such as GPT-3.5 (used in this study), to generate realistic and relevant responses to the proposed scenarios. However, these models may have limitations in understanding the specific context or providing accurate answers in certain situations.

**Technological and Strategic Limitations:** The experimental platform at this time used technologies that limited its capabilities, since GPT-3.5 was used as the main Generative AI engine, which by nature presents certain input size restrictions or limitations in dealing with complex contexts.

Strategies such as (Fine-tuning, Grounding LLMs, Retrieval-Augmented Generation-RAG, ML, Analytics) were not used, which directly contribute to improving the understanding and responsiveness of the model, allowing it to connect more effectively with the real world and generate more relevant and useful answers to the queries presented to it.

**Bias in the Definition of Scenarios:** We understand that the definition of scenarios and questions can introduce bias into the simulation, affecting the results and conclusions obtained. The way scenarios are structured, and questions are formulated can influence the answers generated by language models and, consequently, the insights obtained.

**Validation of Results:** Validating the results obtained in the simulation can be challenging, since they are generated by language models and may not necessarily reflect reality. We use a simplified methodology focusing on devices used in qualitative analysis.

Although there are limitations and challenges to be overcome, the integration of multi-agent systems (SMA) and large language models (LLM) represents a unique opportunity to transform human-machine interaction in modern organizations. By proactively recognizing and addressing these limitations, organizations can fully leverage the potential of these innovative approaches to improve the efficiency, accuracy, and adaptability of their operations.

# 5 Ethical Considerations

It is important to recognize that the use of AI brings with it ethical and regulatory issues, especially related to data protection and transparency in decision-making processes. As organizations explore the possibilities of this technology, it is critical to maintain a balance between the pursuit of innovation and protecting the interests of everyone involved (not exhaustive).

- **Privacy and Data Protection:** The proposed framework aims to enable human participation in conversations and processes that may involve human activity combined with AI.

- **Bias and Fairness:** In the case of using LLMs, it is known that they can exhibit biases, caused by the nature of the technology or the use of training data, which is important in any solutions of this nature to mitigate prejudices, deviations, and other forms of imbalance.

- **Responsibility and Transparency:** The structure may involve multiple agents talking and cooperating, it is important to establish clear responsibility and transparency mechanisms. The platform under development will understand, track, and govern decision making from data selection to the design of agents and metahumans.

# 6 Conclusions and Reflexions

This study explored the revolutionary potential of multi-agent systems (SMA) combined with large language models (LLM) in transforming competition into collaboration considering aspects for modern organizations, we used only a short simulation of a simplified debate between two professionals with different experiences and knowledge, but with convergent objectives, using simplified code, elementary components, and prompt engineering.

We understand that the greatest contribution is the demonstration that by employing specialized multi-agents, developed from LLMs, demonstrates the ability to create diverse solutions and useful meta-humans in different areas of knowledge such as:

- **Information Retrieval:** LLM can quickly gather information from various sources and provide relevant data to support decision-making processes within organizations.

- **Task Automation:** LLM can automate repetitive tasks, such as scheduling meetings, answering common queries, or generating reports, freeing up human resources to focus on more complex and strategic tasks.

- **Customer Support:** LLM can assist in customer support by answering frequently asked questions, troubleshooting issues, and providing personalized recommendations based on customer inquiries.
- **Training and Education:** LLM can facilitate training and education by providing interactive learning experiences, debates, point of views, answering questions, and offering explanations on complex topics.
- **Collaborative Problem-Solving:** LLM can collaborate with humans and other agents to solve complex problems by providing insights, suggestions, and alternative perspectives.
- **Language Translation and Interpretation:** LLM can assist in language translation and interpretation tasks, helping to bridge communication gaps between individuals and organizations from different linguistic backgrounds.
- **Content Creation and Curation:** LLM can generate content for marketing materials, social media posts, and other communication channels, as well as curate relevant content from the web.
- **Decision Support:** LLM can analyze data, generate insights, and provide recommendations to support decision-making processes within organizations.

This study represents an initial step in exploring the potential to transform the way organizations use collaborative tools and resources, especially by combining concepts, theories, principles, and practices related to Multi-Agent Systems (SMA), Artificial Intelligence (AI), Large Models of Language (LLM), Artificial, Archetypes, Personas, meta-humans, design, suggested human behavior and questions intrinsic to each topic that are great in themselves.

Future research can focus on more advanced approaches, considering other industries, other activities, other personas, for example, performing validations and more robust analysis of results, aiming to create more adaptable and effective systems to deal with organizational challenges in a constantly changing environment. evolution.

The platform mentioned in this study is evolving and new developments are expected to become useful to humans in their various areas of knowledge, enabling AI to transform competition into real collaboration, a window that opens to the resilient future with the use competent and creative Artificial Intelligence.